\documentclass[letterpaper]{article} 
\usepackage{aaai23}  
\usepackage{times}  
\usepackage{helvet}  
\usepackage{courier}  
\usepackage[hyphens]{url}  
\usepackage{graphicx} 
\usepackage{natbib}  
\usepackage{caption} 
\usepackage{algorithm}
\usepackage{algorithmic}
\usepackage{amssymb}
\usepackage{soul}
\usepackage{xspace}
\usepackage{mathrsfs}
\usepackage{amsmath}
\usepackage{amsthm}
\usepackage{multirow}
\usepackage{varioref}
\usepackage{array}
\usepackage{verbatim}
\usepackage{makecell}
\usepackage{bm}

\newcommand{\baby}{VWB-\textsc{cmr}\xspace}

\newtheorem{theorem}{Theorem}
\newtheorem{definition}{Definition}
\usepackage{newfloat}
\usepackage{listings}
\DeclareCaptionStyle{ruled}{labelfont=normalfont,labelsep=colon,strut=off} 
\floatstyle{ruled}
\newfloat{listing}{tb}{lst}{}
\floatname{listing}{Listing}
\title{Variational Wasserstein Barycenters with $c$-Cyclical Monotonicity Regularization}
\author{
    Jinjin Chi\textsuperscript{\rm 1,\rm 2},
    Zhiyao Yang\textsuperscript{\rm 1,\rm 2},
    Ximing Li\textsuperscript{\rm 1,\rm 2} \thanks{Corresponding author},
    Jihong Ouyang\textsuperscript{\rm 1,\rm 2},
    Renchu Guan\textsuperscript{\rm 1,\rm 2}\\
}
\affiliations{
    \textsuperscript{\rm 1} College of Computer Science and Technology, Jilin University, China\\
    \textsuperscript{\rm 2} Key Laboratory of Symbolic Computation and Knowledge Engineering of Ministry of Education, China

    chijinjin616@gmail.com, yangzy9529@gmail.com, liximing86@gmail.com, ouyj@jlu.edu.cn, guanrenchu@jlu.edu.cn
}

\usepackage{bibentry}

\begin{document}

\maketitle

\begin{abstract}
Wasserstein barycenter, built on the theory of Optimal {T}ransport ({OT}), provides a powerful framework to aggregate probability distributions, and it has increasingly attracted great attention within the machine learning community. However, it is often intractable to precisely compute, especially for high dimensional and continuous settings. To alleviate this problem, we develop a novel regularization by using the fact that $c$-cyclical monotonicity is often necessary and sufficient conditions for optimality in OT problems, and incorporate it into the dual formulation of Wasserstein barycenters. For efficient computations, we adopt a variational distribution as the approximation of the true continuous barycenter, so as to frame the Wasserstein barycenters problem as an optimization problem with respect to variational parameters. Upon those ideas, we propose a novel end-to-end continuous approximation method, namely Variational Wasserstein Barycenters with $c$-Cyclical Monotonicity Regularization (\baby), given sample access to the input distributions. We show theoretical convergence analysis and demonstrate the superior performance of \baby on synthetic data and real applications of subset posterior aggregation.
\end{abstract}

\section{Introduction}
Summarizing, combining, and comparing probability distributions defined on a metric are fundamental yet vital tasks in machine learning, statistics, and computer science. For instance, in Bayesian inference, given massive data one may perform posterior inference in parallel on different machines with data subsets, and then aggregate subset posterior distributions via their \emph{barycenter} as an approximation to the full data posterior \citep{2015wasp}. Besides, the barycenter of a collection of distributions has been widely studied in various real applications, such as image processing \citep{rabin2011} and clustering \citep{ho2019}.

\begin{figure}[!tbh]
\includegraphics[width=0.5\textwidth]{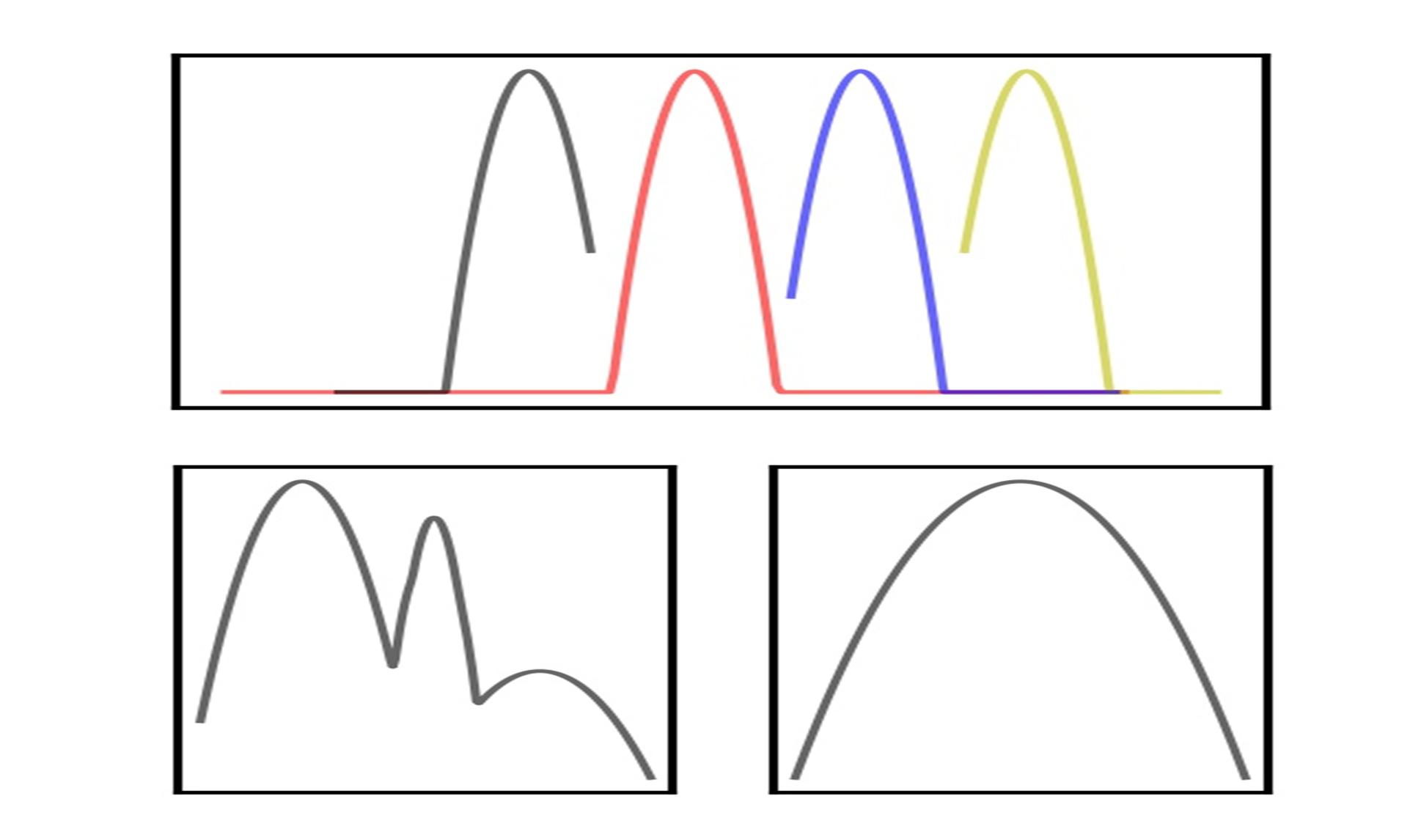}
\centering
\caption{Top: four distributions. Bottom left: Euclidean average of distributions. Bottom right: Wasserstein barycenter. }
\label{Fig1}
\end{figure}

The theory of \textbf{O}ptimal \textbf{T}ransport (\textbf{OT}) \citep{monge1781,kantorovich1942,villani2008,le2021robust,backhoff2022} provides a powerful framework to carry out such aggregation. OT equips the space of distributions with a distance metric known as the \emph{Wasserstein distance}, which has gained substantial popularity in different fields. The \emph{barycenter} of multiple given probability distributions under \emph{Wasserstein distance} is defined as \textbf{Wasserstein barycenter}, i.e., a distribution minimizing the sum of Wasserstein distances to all input distributions. Due to geometric properties, the Wasserstein barycenter can better capture the underlying geometric structure than the barycenter with respect to other popular distances, e.g., Euclidean distance, see Figure \ref{Fig1}. Consequently, it has a broad range of applications in text mixing \citep{rabin2011}, imaging \citep{bonneel2016}, and model ensemble \citep{dognin2019}.

Despite its impressive performance, it is often intractable to precisely compute Wasserstein barycenters, especially for high dimensional and continuous settings \citep{peyre2019, Korotin2021bench,daaloul2021}. The existing studies \citep{agueh2011,alvarez2016,dvurechenskii2018,ge2019,lin2020} on Wasserstein barycenter of continuous distributions require discretization of all input distributions or the barycenter itself, so they scale poorly to high-dimensional settings \citep{staib2017,claici2018,dvurechenskii2018,izzo2021dimensionality}. Moreover, the discretization techniques are commonly undesirable since they lack the inherently continuous nature of data distributions and the ability of generating new samples if needed. To the best of our knowledge, very few works attempt to operate directly on continuous distributions for computing a barycenter. In this work, we further contribute on this challenging topic of continuous barycenters.
\begin{table}[t]
\label{Table1}
\renewcommand\arraystretch{1.2}
\newcommand{\tabincell}[2]{\begin{tabular}{@{}#1@{}}#2\end{tabular}}
\begin{center}
\begin{small}
\begin{tabular}{p{0.33\columnwidth}c p{0.18\columnwidth} c}
\Xhline{1.2pt}
Literature & \tabincell{c}{General \\cost} & \tabincell{c}{Barycenter} &   End-to-end  \\
\hline
\cite{li2020} & $\surd$ & Fixed prior & $\times$ \\
\cite{fan2020}& $\times$ & {Generative network} & $\times$ \\
\cite{korotin2021}& $\times$ &Fixed prior & $\times$\\
\cite{korotin2022}& $\times$ &{Generative network}& $\times$\\
Our method & $\surd$ & Variational distribution & $\surd$\\
\Xhline{1.2pt}

\end{tabular}
\end{small}
\end{center}
\caption{Summary of existing Wasserstein barycenter methods.}
\end{table}

\paragraph{Our contributions.} We propose a computationally efficient and straightforward method for estimating continuous Wasserstein barycenters, given sample access to the input distributions. The key idea is to introduce a novel $c$-cyclical monotonicity regularization
following the fact that $c$-cyclical monotonicity is often necessary and sufficient conditions for optimality in OT problems. 
We incorporate this regularization into the dual formulation of the Wasserstein barycenter problem, which guarantees the learning of optimal dual potentials, so as to obtain a good approximation of the barycenter. Our method is end-to-end without recovering the barycenters from the dual solution. This is made possible by introducing a family of continuous distributions, namely variational distribution, and using the closest one as the approximation of the barycenter. We thereby obtain a tractable objective with respect to variational parameters, which can be efficiently solved through stochastic optimization. We provide theoretical analysis on convergence as well as empirical evidence of the effectiveness of our method on both synthetic data and real applications of subset posterior aggregation.
\paragraph{Related work.}
The notion of the Wasserstein barycenter was first introduced by \citep{agueh2011} and then investigated by many works via the original Wasserstein distance as well as its regularization. Most of them mainly solve discrete Wasserstein barycenter problem, namely, averaging some discrete distributions, via linear programs (or some equivalent problem) \citep{anderes2016,cuturi2018,yang2018,ge2019} or regularized projection-based methods \citep{cuturi2014fast,benamou2015,cuturi2016,ye2017fast,janati2020}. However, these methods can not solve the continuous Wasserstein barycenter problem. To address this issue, \citep{staib2017,claici2018,dvurechenskii2018} assume the barycenter to be a finite set of points and rely on semi-discrete OT algorithms to compute the barycenter. 

The continuous approximation methods for Wasserstein barycenters remain unexplored until recently \citep{li2020,fan2020,korotin2021,korotin2022}. Existing methods are based on dual potentials optimization or fixed point approaches. However, these methods are not \emph{end-to-end} which consist of two sequential steps. They require recovering barycenters via an additional estimation, e.g., computing gradients of the potentials as push-forward maps or estimating Monge maps, which is itself a complex problem increasing the computational burden of the original problem. Closely related to this paper are the recent works \citep{li2020,fan2020,korotin2021} that rely on the dual formulation of the Wasserstein barycenter and represent the dual potentials with neural networks. In particular, the method in \citep{li2020} assumes a \emph{fixed prior} as the proxy of the barycenter from the beginning, leading to inaccurate approximations especially for high-dimensional settings \citep{korotin2022}. For the specific ground cost, i.e., \emph{quadratic cost}, \citep{fan2020,korotin2021} compute the barycenters under Wasserstein-2 distance. The method in \citep{fan2020} employs the \emph{generative network} to represent a barycenter, which suffers from the usual limitations of generative networks such as mode collapse. Although this method does not involve regularization terms, it ends up with a challenging min-max-min problem. Another well-known method in \citep{korotin2021} constructs a new regularized formulation by adding a congruence regularization to ensure that optimal potential functions are consistent with the true barycenter. However, the congruence regularization requires the selection of a \emph{prior} distribution that is bounded below by the barycenter, which is a non-trivial task. On the other hand, \citep{korotin2022} proposes an iterative algorithm for approximating the Wasserstein-2 barycenters based on the fixed point approach, where a \emph{generative network} is used to parameterize the evolving barycenter. However, it is challenging to guarantee convergence to the true barycenter. Besides, \citep{cohen2020} proposes a generic algorithm to compute barycenters with respect to arbitrary discrepancies, which also parameterizes the barycenter by using a generative network. In contrast, our method is end-to-end and suitable for a general ground cost distance. Remarkably, we use a parameterized variational distribution as the approximation of the barycenter, allowing better flexibility. Table 1 summarizes the existing studies on continuous Wasserstein barycenter and shows our contributions.

\section{Background and Preliminaries}
In this section, we describe the Wasserstein barycenter problem, which involves all Wasserstein distances from one to many distributions.
\begin{definition}[Wasserstein distance]
 Let $\mathcal{X}$ and $\mathcal{Y}$ be arbitrary spaces equipped with a ground cost $c(x,y)=d(x,y)^{p}$~($d(\cdot,\cdot)$ is a distance). Given two continuous probability distributions $\mu(x)$ and $\nu(y)$, for $p\in[1,\infty)$, the Wasserstein distance between $\mu$ and $\nu$ is defined as follows \citep{villani2008}:
\begin{equation}
\label{wd}
 W_{p}^{p}(\mu,\nu)\buildrel \Delta \over=\mathop{\mathbf{inf}}\limits_{\pi  \in \Pi(\mu,\nu)} \:\: \int_{\mathcal{X}} \int_{\mathcal{Y}} {\pi(x,y)c(x,y)dxdy},
\end{equation}
where $\Pi (\mu,\nu)$ is the set of all joint distributions on $\mathcal {X} \times \mathcal {Y}$ with prescribed marginals $\mu(x)$ and $\nu(y)$. We also call an admissible $\pi$ a transport plan. 
\end{definition}

The primal problem (\ref{wd}) admits an equivalent dual form \citep{villani2008}:
\begin{equation}
\label{wd1}
\mathop{\mathbf{sup}}\limits_{\phi \oplus \psi \le c} \:\: \int_{\mathcal{X}}\phi(x)\mu(x)dx+ \int_{\mathcal{Y}} {\psi(y)\nu(y)dy},
\end{equation}
where $\phi$ and $\psi$ are dual potentials, and $(\phi \oplus \psi)(x,y)  \buildrel \Delta \over = \phi(x)+\psi(y)$. The constraint $\phi \oplus \psi \le c$ means $\phi(x) + \psi(y) \le c(x,y)$ for all $(x,y)$.

Directly solving (1) and (2) is challenging, since the resulting linear problem can be too costly. In order to speed up the computation, regularized OT is introduced by \citep{cuturi2013}. Here, we consider the entropy and quadratic regularization:
\begin{eqnarray}
\label{wd2}
\!\!\!\!\!\!\!\!\!\!\!\!\!\!\!\!\!\!&&\mathop{\mathbf{sup}}\limits_{\phi,\psi } \:\: \int_{\mathcal{X}}\phi(x)\mu(x)dx+\int_{\mathcal{Y}}\psi(y)\nu(y)dy\nonumber\\
\!\!\!\!\!\!\!\!\!\!\!\!\!\!\!\!\!\!&&~~~~~~+\int_{\mathcal X}\int_{\mathcal Y}F(\phi (x) + \psi (y) - c(x,y))\mu(x)\nu(y)dxdy,
\end{eqnarray}
where
\begin{equation}
\label{wd_reg}
\forall t \in \mathbb{R},\:\:\: F(t) =\left\{ {\begin{array}{*{1}{l}}
{-\varepsilon {e^{{\textstyle{t \over \varepsilon }}}}\:\:\:\:\:\:\:\:\:\:\:\:\:\:(\rm{entropy\:\: reg.} )}  \\
{-\frac{1}{{2\varepsilon }}{{(t_{+})}^2}\:\:\:\:\:\:(\rm{quadratic}\:\: \rm{reg.})}\\
\end{array}} \right.
\end{equation}

\begin{definition}[Wasserstein barycenter]
The Wasserstein barycenter of $N~(N \ge 2)$ continuous probability distributions $\mu_n$, $n=1,2,\ldots,N$ with weights $\alpha_n$ $(\alpha_n\textgreater 0,\sum\nolimits_{n = 1}^N \alpha_n =1)$, is a solution of the following functional minimization problem \citep{agueh2011} :
\begin{equation}
\label{wb}
\!\!\!\!\!\!\!\!\!\!\!\!\!\mathop {\mathbf{inf } }\limits_\nu  \sum\limits_{n = 1}^N {\alpha_n W_p^p(\mu_n ,{\nu})}
\end{equation}
\end{definition}

\paragraph{Regularized Wasserstein barycenter dual.}
Since the primal objective (\ref{wb}) is hard to compute, \citep{li2020} defines a new dual version of regularized Wasserstein barycenter problem by using regularized Wasserstein distance (\ref{wd2}), which is a more popular and efficient alternative:
\begin{eqnarray}
\label{rwb}
\!\!\!\!\!\!\!\!\!\!\!&&\mathop {\mathbf{inf } }\limits_\nu \mathop{\mathbf{sup}}\limits_{\phi,\psi \atop {\sum\limits_{n = 1}^N \alpha_n \psi_n=0}} \sum\limits_{n= 1}^N\alpha_n\Bigg[ {\int_{\mathcal X}}\phi_n(x_n) \mu_n(x_n) dx_n + \\
\!\!\!\!\!\!\!\!\!\!\!&&\int_{\mathcal X}\int_{\mathcal Y}F({\phi_n(x_n)+ {\psi_n(y)}-c(x_n,y)})\mu_n(x_n)\nu(y) dx_n dy \Bigg]\nonumber
\end{eqnarray}
where $\phi_n$ and $\psi_n$ are dual potentials, and $F$ refers to either entropy or quadratic regularization. We notice that $\nu$ is the true barycenter, not a prior like in \citep{li2020}. We rely on this formulation in the next section to propose a new regularized objective.

\section{Our Method}
Our goal in this paper is to compute the Wasserstein barycenter for a given set of continuous distributions $\left\{\mu_1,\ldots,\mu_N\right\}$. To effectively solve this problem, we present a novel end-to-end method, namely \textbf{V}ariational \textbf{W}asserstein \textbf{B}arycenter with $c$-\textbf{C}yclical \textbf{M}onotonicity \textbf{R}egularization (\textbf{\baby}). Our method is well-suited for the case where the analytic forms of input distributions are not available. Since we only have access to independent samples which can be drawn from input distributions or are provided by machine learning applications.

\subsection{Imposing the $c$-Cyclical Monotone Condition}
Inspired by \citep{li2020}, we rely on a regularized formulation to compute the Wasserstein barycenter. However, (\ref{rwb}) cannot provide precise approximations in high dimensions \citep{korotin2021}. To address this issue, we enforce $c$-cyclical monotone condition via a regularization.

 A set $\Gamma \subseteq \mathcal X \times \mathcal Y$ is said $c$-\emph{cyclical monotone}, if for any $k \in \mathbb{N}$ $(k\ge 2)$, any permutation $\sigma \in P(k)$ and any finite pairs $(x^1,y^1)$,...,$(x^k,y^k) \in \Gamma$, one can have
\begin{equation}
\sum\limits_{i=1}^k c(x^i,y^i) \le \sum\limits_{i=1}^k c(x^i,y^{\sigma(i)}),
\end{equation}
where $P(k)$ is the set of all permutations of $\{1,...,k\}$. For a finite set $\Gamma$, $c$-cyclical monotonicity means the points of origin $x^i$ and destination $y^i$ related by $(x^i,y^i)$ have been paired so as to minimize the total transportation cost $\sum\nolimits_{\Gamma}c(x,y)$. Otherwise, it would be more efficient to move mass from all $x^i$ to $y^i$.

Importantly, $c$-cyclical monotonicity is a necessary condition for optimality in OT \citep{gangbo1996}, and is a sufficient condition on a weak assumption \citep{pratelli2008}.
\begin{theorem} [Necessary and sufficient optimality conditions]
Assume a cost function $c: \mathcal X \times\mathcal Y \to \mathbb{R} \cup +\infty$ is continuous. Then a transport plan $\pi$ for the Wasserstein distance is optimal (i.e., it is a solution of problem (1)) if and only if the support of $\pi$, i.e., $\Gamma$, is $c$-cyclical monotone.
\end{theorem}

This theorem can be proved by contradiction. The detailed discussion can be found in \citep{pratelli2008}.

In order to enforce $c$-cyclical monotonicity, we define
\begin{equation*}
    I(x,y)\buildrel \Delta \over=\mathop{\mathbf{sup}}\limits_{\sigma \in P(k)}  \sum\limits_{i=1}^k c(x^i,y^i) - \sum\limits_{i=1}^k c(x^i,y^{\sigma(i)})
\end{equation*}
Then $I: \mathcal{X} \times \mathcal{Y} \to [0,\infty]$. For all $(x,y) \in \Gamma$, the optimality of $\pi$ implies that $I=0$ since $\Gamma$ is $c$-cyclically monotone. Combing the constraint of dual formulation $\phi(x^i)+\psi(y^i)\le c(x^i,y^i)$, we want to maximize $\sum\nolimits_{i=1}^k\phi(x^i)+\psi(y^i)- c(x^i,y^{\sigma(i)})$. The supremum value of this difference is zero when $(x^i,y^i) \in \Gamma$. But outside $\Gamma$, we expect this supremum value is greater than zero. Hence we consider an additional convex regularization:
\begin{equation}
\label{cr}
F(\sum\limits_{i=1}^k c(x^i,y^{\sigma(i)})-\phi(x^i)-\psi(y^i)),
\end{equation}
where $F$ refers to either entropy or quadratic regularization defined in (\ref{wd_reg}). As a result, this regularization can help to find the optimal dual potentials, so as to make more precise approximations of barycenters, especially for high-dimensional settings, see more details in experiments.

Plugging (\ref{cr}) into (\ref{rwb}), we obtain a new regularized objective of the Wasserstein barycenter problem:
\begin{eqnarray}
\label{obj}
\!\!\!\!\!\!\!\!\!\!\!\!\!\!\!\!\!& &\mathop {\mathbf{inf } }\limits_\nu\mathop{\mathbf{sup}}\limits_{\phi,\psi \atop {\sum\limits_{n = 1}^N \alpha_n \psi_n=0}} \sum\limits_{ n= 1}^N\alpha_n\Bigg[ {\int_{\mathcal X}}\phi_n(x_n) \mu_n(x_n) dx_n+ \nonumber\\
\!\!\!\!\!\!\!\!\!\!\!\!\!\!\!\!\!& &~~~~~~~~~~~~~\int_{\mathcal X}\int_{\mathcal Y} R(\phi_n(x_n),\psi_n(y))\mu_n(x_n)\nu(y) dx_n dy \Bigg],
\end{eqnarray}
where
\begin{eqnarray}
\label{reg}
\!\!\!\!\!\!\!\!\!\!\!\!\!\!\!\!& &R(\phi_n(x_n),\psi_n(y))\buildrel \Delta \over =F({\phi_n(x_n)+ \psi_n(y)-c(x_n,y)})+\nonumber\\
\!\!\!\!\!\!\!\!\!\!\!\!\!\!\!\!& &~~~~~~~~~~~~~~~~~~~~~F(\sum\limits_{l=1}^k c(x^l_n,y^{\sigma(l)})-\phi_n(x^l_n)-\psi_n(y^{l}))
\end{eqnarray}

For a fixed barycenter $\nu$, (\ref{obj}) is concave with respect to $\phi_n$ and $\psi_n$ which can be maximized through stochastic gradient methods by using deep neural networks for parameterizing $\phi_n$ and $\psi_n$.

\paragraph{Discussion.} Though the $c$-cyclical monotone condition is true for any permutation $\sigma$, it is enough to check this condition by using one permutation. In practice, we have examined more permutations and found that using a single permutation can make good performances in most cases. There are cases where using two permutations is slightly better than using just one. However, the performance does not get significantly better as the permutation increases, and is even worse than using one or two permutations. We believe that more regularization terms can introduce bias. In our experiments, we only report the results of using a single permutation $\sum\nolimits_{l=1}^k c(x^l,y^{k+1-l})$ in $c$-cyclical monotonicity regularization, and set $k$ equal to the number of Monte Carlo samples as indicated in \textbf{\emph{Algorithm 1}}. 

\subsection{Introducing a Variational Distribution}
In the context of estimating barycenters of continuous distributions, we propose to represent the barycenter by using a variational distribution $\nu'(\cdot|\lambda)$ with parameters $\lambda$. If we know the input distributions are in the same family, e.g., Gaussian, we can set the variational distribution $\nu'$ to be a Gaussian with unknown mean and covariance. Otherwise we may consider $\nu'=\mathop \Pi \nolimits_n^{N} \nu'_n(\cdot|\lambda_n)$, where the type of $\nu'_n$ coincides with each input distribution $\mu_n$. In the case where a fixed set of samples is provided and no information about the barycenter is known beforehand, we can assume that the variational distribution is a Gaussian mixture, which has the ability to approximate any functions \citep{anzai2012}.

Having specified a variational distribution $\nu'(y|\lambda)$, we transform the barycenter estimation problem into an optimization problem, where the parameters $\lambda$ to be optimized adjust a variational “proxy” distribution to be similar to the barycenter. We rewrite (\ref{obj}) with variational distribution $\nu'(y|\lambda)$ and replace each $\psi_n$ with $\psi_n-\sum\nolimits_{j = 1}^N\alpha_j \psi_j$ to obtain an unconstrained version as follows:
\begin{eqnarray}
\label{obj1}
\!\!\!\!\!\!\!\!\!\!\!\!\!\!\!\!\!\!& &\mathop {\mathbf{inf } }\limits_\lambda\mathop{\mathbf{sup}}\limits_{\phi,\psi}\mathcal{L(\lambda,\phi,\psi)}\buildrel\Delta\over=\sum\limits_{ n= 1}^N\alpha_n\Bigg[ {\int_{\mathcal X}}\phi_n(x_n) \mu_n(x_n) dx_n+ \nonumber\\
\!\!\!\!\!\!\!\!\!\!\!\!\!\!\!\!\!\!& &~~~~~~~~\int_{\mathcal X}\int_{\mathcal Y} R(\phi_n(x_n),\psi_n(y))\mu_n(x_n)\nu'(y|\lambda) dx_n dy \Bigg],
\end{eqnarray}
where
\begin{eqnarray}
\label{reg1}
\!\!\!\!\!\!\!\!\!\!\!\!\!\!\!\!\!\!\!& &R(\phi_n(x_n),\psi_n(y)) =\nonumber\\
\!\!\!\!\!\!\!\!\!\!\!\!\!\!\!\!\!\!\!& &F(\sum\limits_{l=1}^k c(x^l_n,y^{\sigma(l)})-\phi_n(x^l_n)-\psi_n(y^{l})+\sum\limits_{j = 1}^N\alpha_j \psi_j(y^l))\nonumber\\
\!\!\!\!\!\!\!\!\!\!\!\!\!\!\!\!\!\!\!& &+F({\phi_n(x_n)+ \psi_n(y)-\sum\limits_{j = 1}^N\alpha_j \psi_j(y)-c(x_n,y)})
\end{eqnarray}

Thus we obtain a tractable objective with respect to variational parameters. We can update variational parameters through stochastic gradient methods.
\paragraph{The noisy gradients of $\mathcal L$.} 
To optimize the objective $\mathcal L$ with gradient-based methods, we need to develop an unbiased estimator of its gradients which can be computed using Monte Carlo samples. To do this, we derive gradients of $\mathcal{L}$ with respect to $\lambda$ as an expectation form:
\begin{eqnarray}
\label{dev_rd}
\!\!\!\!\!\!\!\!\!\!\!\!\!& &{\nabla _\lambda }\mathcal{L} \nonumber\\
\!\!\!\!\!\!\!\!\!\!\!\!\!& &={\nabla _\lambda }\sum\limits_{ n= 1}^N\alpha_n
 \int_{\mathcal X}\int_{\mathcal Y} R(\phi_n(x_n),\psi_n(y))\mu_n(x_n)\nu'(y|\lambda) dx_n dy\nonumber\\
\!\!\!\!\!\!\!\!\!\!\!\!\!& &=\sum\limits_{ n= 1}^N\alpha_n
 \int_{\mathcal X}\int_{\mathcal Y} {\nabla _\lambda }R(\phi_n(x_n),\psi_n(y))\mu_n(x_n)\nu'(y|\lambda) dx_n dy \nonumber\\
\!\!\!\!\!\!\!\!\!\!\!\!\!& &=\sum\limits_{ n= 1}^N\alpha_n
 \mathbb{E}_{\mu_n\nu'} [{\nabla _\lambda }\log\nu'(y|\lambda)R(\phi_n(x_n),\psi_n(y))]
\end{eqnarray}
where ${\nabla_\lambda}\log\nu'(y|\lambda)$ is called the score function \citep{cox1979}, and it can be calculated using automatic differentiation tools \citep{stan2015}. Note that ${\nabla _\lambda }[\nu'(y|\lambda)]={\nabla _\lambda }[\log\nu'(y|\lambda)]\nu'(y|\lambda)$.

With this equation in hand, we can directly compute the gradient using Monte Carlo samples drawn from $\mu(x)\nu'(y|\lambda)$:
\begin{eqnarray*}
\label{sr_lambda}
\!\!\!\!\!\!\!\!\!\!\!\!\!& & {\nabla _\lambda }{\mathcal L}  \approx \frac{1}{S}\sum\limits_{n = 1}^N\sum\limits_{s = 1}^S\alpha_n{\nabla_\lambda}\log\nu'(y^{(s)}|\lambda) R(\phi_n(x_n^{(s)}),\psi_n(y^{(s)})\nonumber\\
\!\!\!\!\!\!\!\!\!\!\!\!\!&&~~~~~~~~~~~~~~~~~~~~  
\end{eqnarray*}
where $S$ is the number of Monte Carlo samples. Then, at each iteration $t$, the parameter of interest $\lambda$ can be updated as follows:
\begin{equation*}
\label{opt}
\lambda_t  \leftarrow \lambda_{t-1}  - {\rho_t}{\nabla _\lambda }{\mathcal L},
\end{equation*}
where $\rho_t$ is the learning rate.

We notice that the optimization over $\nu$ is not convex. However, it guarantees to converge to a local optimum, if the learning rate satisfies the Robbins-Monro condition \citep{robbins1951}:
\begin{equation*}
\sum\nolimits_{t = 1}^\infty  {{\rho_t } = \infty},\:\: \sum\nolimits_{t = 1}^\infty  {\rho_t ^2 < \infty} \nonumber
\end{equation*}

\begin{algorithm}[tb]
\caption{Optimization of \baby}
\label{althm1}
\begin{algorithmic}[1] 
\STATE \textbf{Input}: Continuous distributions $\left\{\mu_n \right\}_{n=1}^N$ with weight $\alpha_n$; cost function $c$; batch size $S$; regularization $F$; learning rate $\rho$; network gradient update function \emph{BackWard}.\\
\STATE \textbf{While} \emph{not converged} do
\STATE $\quad$  $\forall n\in\{1,...,N\}$: sample $x_n$ from $\mu_n$
\STATE $\quad$ sample $y$ from $\nu'(y|\lambda)$ and obtain a permutation $y^{\sigma}$
\STATE $\quad$ $\overline{\psi}_y \leftarrow \sum\nolimits_{j=1}^{N}\alpha_j \psi_j(y)$
\STATE $\quad$ \textbf{For} $n=1$ to $N$ do
\STATE $\quad$ ~~ $R_n^1 \leftarrow F(\sum\limits_{l=1}^S c(x^l_n,y^{\sigma(l)})-\phi_n(x^l_n)-\psi_n(y^{l})+\overline{\psi}_{y^l})$
\STATE $\quad$ ~~ $R_n^2 \leftarrow F({\phi_n(x_n)+ {\psi_n(y)-\overline{\psi}_y}-c(x_n,y)})$
\STATE $\quad$ ~~ $R_n \leftarrow R_n^1+R_n^2$
\STATE $\quad$ \textbf{End For}
\STATE $\quad$ $ \mathcal{L} \leftarrow \sum\nolimits_{ n= 1}^N\alpha_n(\phi_n(x_n) + R_n)$
\STATE $\quad$ \textbf{For} $n=1$ to $N$ do \textbf{in parallel}
\STATE $\quad$ $\quad$ \emph{BackWard($\mathcal{L},\phi_n$)}; $\quad$ \emph{BackWard($\mathcal{L},\psi_n$)}
\STATE $\quad$ \textbf{End For}
\STATE $\quad$ $f \leftarrow  \sum\nolimits_{n=1}^{N}\alpha_n{\nabla _\lambda } \log\nu'(y|\lambda)R_n$
\STATE $\quad$ $h \leftarrow  {\nabla_\lambda}\log\nu'(y|\lambda)$
\STATE $\quad$ $ {a^ * }\leftarrow {{{\rm{Cov}}(f,h)} \mathord{\left/ {\vphantom {{{\rm{Cov}}(f,h)} {{\rm{Var}}(h)}}} \right. \kern-\nulldelimiterspace} {{\rm{Var}}(h)}} $
\STATE $\quad$ update $\lambda$ $\leftarrow$ $\lambda-\rho(f-a^*h)$
\STATE \textbf{End While}
\STATE \textbf{Output}: The continuous Wasserstein barycenter $\nu'$ of $\left\{\mu_n \right\}_{n=1}^N$.
\end{algorithmic}
\end{algorithm}

\paragraph{Variance reduction.}
The noisy gradients formed by using Monte Carlo samples often can be too large to be useful. In practice, the high variance would lead to slow convergence and even worse performance \citep{PaisleyBJ12}.

To alleviate this, we use the control variate to reduce variance \citep{ross2002, PaisleyBJ12}, which is a family of functions with equivalent expectations. The basic idea is to replace the target function with another function that has the same expectation but a smaller variance. For example, when we compute $\mathbb{E}[f]$ with Monte Carlo samples, we use the empirical average of $\hat f$ where $\hat f$ is chosen so $\mathbb{E}[f]=\mathbb{E}[\hat f]$ and ${\rm{Var}}[f] \textgreater {\rm{Var}}[\hat f]$. Next, we describe how to reduce the variance via easy-to-implement control variates.

Define $\hat f$ to be
\begin{equation*}
\hat f(z)\buildrel \Delta \over =f(z)-a(h(z)-\mathbb E[h(z)]),
\end{equation*}
where $h(z)$ is a function with a finite first moment, and $a$ is a scalar. Note that $\mathbb{E}[f]=\mathbb{E}[\hat f]$ as required.
The variance of $\hat f$ is:
\begin{equation*}
{\rm{Var}}(\hat f)={\rm{Var}}(f)+a^2{\rm{Var}}(h)-2a{\rm{Cov}}(f,h)
\end{equation*}
This equation implies that a good control variate $\hat f$ with a smaller variance will have high covariance with the function $f$. Given a function $h$, taking the derivative of ${\rm{Var}}(\hat f)$ with respect to $a$ and setting it equal to zero, one can get the optimal scaling,
\begin{equation*}
   {a^ * } = {{{\rm{Cov}}(f,h)} \mathord{\left/ {\vphantom {{{\rm{Cov}}(f,h)} {{\rm{Var}}(h)}}} \right. \kern-\nulldelimiterspace} {{\rm{Var}}(h)}},
\end{equation*}
 which can be estimated with the ratio of empirical covariance and variance using Monte Carlo samples.

Inspired by this, in our method we choose $h$ to be the score function of the variational distribution, i.e., ${\nabla_\lambda}\log\nu'(y|\lambda)$ which always has expectation zero, and re-compute the gradient of $\mathcal L$ with respect to $\lambda$ using a new Monte Carlo method:
\begin{eqnarray}
\label{sr_lambda1}
\!\!\!\!\!\!\!\!\!\!\!\!\!& & {\nabla _\lambda }{\mathcal L}  \approx \nonumber\\
\!\!\!\!\!\!\!\!\!\!\!\!\!& &\frac{1}{S}\sum\limits_{n = 1}^N\sum\limits_{s = 1}^S\alpha_n{\nabla_\lambda}\log\nu'(y^{(s)}|\lambda) (R(\phi_n(x_n^{(s)}),\psi_n(y^{(s)})-a^*)\nonumber\\
\!\!\!\!\!\!\!\!\!\!\!\!\!&&~~~~~~~~~~~~~~~~~~~~~~~\quad\quad\quad\quad  x_n^{(s)} \sim \mu_n(x_n),y^{(s)}\sim \nu'(y|\lambda)
\end{eqnarray}

Finally, we summarize the full algorithm of \baby in \textbf{\emph{Algorithm 1}}.

\paragraph{Computation complexity.} For our method, as well as the methods proposed in \citep{li2020,korotin2021,fan2020,korotin2022}, estimating Wasserstein barycenters includes repeated computation on $N$ input distributions. Therefore, the computational complexity per iteration scales with $O(NSP)$ where $P$ is the size of the networks. Besides, these previous methods require recovering the barycenter by an additional calculation, which is not an easy task scaling with square level complexity at least. However, our method is end-to-end and can easily be parallelized, where $N$ dual variable pairs $(\phi_n,\psi_n)$ can be computed in a fully parallel manner.
\subsection{Theoretical Results}
In this section, we discuss the theoretical guarantee for the convergence of our method.

The work \citep{agueh2011} shows if at least one of the distributions $\mu_n$ is absolutely continuous, then the Wasserstein barycenter $\nu$ is unique. Moreover, $\nu$ is also absolutely continuous. This analysis ensures the existence and uniqueness of the Wasserstein barycenter for continuous distributions.

\begin{theorem}[Convergence]
Let $\mu_1,...,\mu_N$ be continuous distributions with respect to the Lebesgue measure. Assume a cost function $c : \mathcal X \times\mathcal Y \to \mathbb{R} \cup +\infty$ is continuous. If $\{\phi_n,\psi_n\}_{n=1}^{N}$ are the optimal dual potentials in (\ref{obj}), then each $\{\phi_n,\psi_n\}$ is a solution to the regularized dual formulation (\ref{wd2}). Let $\nu^{\varepsilon}$ the solution of regularized Wasserstein barycenter problem (\ref{obj}), and let $\varepsilon$ converge to 0 sufficiently fast. Then, $\nu^{\varepsilon}$ converges weakly to the solution $\nu$ of the Wasserstein barycenter problem (\ref{wb}).
\end{theorem}

We include the proof of Theorem 2 in the supplementary document.
  
\section{Experiments}
In this section, we demonstrate the effectiveness of \baby on both synthetic and real data.

\subsection{Experimental Setup}
Our aim is to examine whether \baby can accurately approximate the continuous barycenter. To this end, we compute the Wasserstein barycenter with the squared Euclidean distance as the cost function, i.e., $c(x,y)=\left\|x-y\right\|_{2}^{2}$, $x,y \in \mathbb{R}^D$. In all experiments, we use equal weights for input distributions, i.e., $\alpha_n=\frac{1}{N}$ for all $n=1,\ldots,N$. Note that the proposed method is not limited to Euclidean distance and equal weights, it can be applied to a more general setting.

We compare \baby against the following state of the art methods: (i) continuous regularized Wasserstein barycenter (CRWB) \footnote{https://github.com/lingxiaoli94/CWB} \citep{li2020}; (ii) continuous Wasserstein barycenter without minimax optimization (CWB) \footnote{https://github.com/iamalexkorotin/Wasserstein2Barycenters} \citep{korotin2021}; (iii) scalable computations of Wasserstein barycenter via input convex neural networks (SCWB) \footnote{https://github.com/sbyebss/Scalable-Wasserstein-Barycenter}\citep{fan2020}. These methods recover barycenter through the gradient-based method. Besides, we consider quadratic regularization in CRWB, since it performs better than entropy regularization.

In our method, we use Adam method to adjust the learning rate, where parameters $\beta_1$=0.9, $\beta_2$=0.999 and $\alpha$=0.001. For the training, the number of iterations and batch size are 20000 and 1024, respectively. The dual potentials $ \left \{ \phi_n,\psi_n \right \}^{N}_{n=1} $ are parameterized as neural networks with two fully-connected hidden layers $(D \to 128 \to 256\to D)$ using ReLU activations. We consider quadratic regularization and report the average results of 5 independent runs for all experiments. The choice of regularization parameter $\varepsilon$ depends on the examples. See the supplementary document for more details and empirical results. 

\subsection{Learning the Wasserstein Barycenter in 2D}

\begin{figure}[hb]
\includegraphics[width=0.5\textwidth,height=0.28\textwidth]{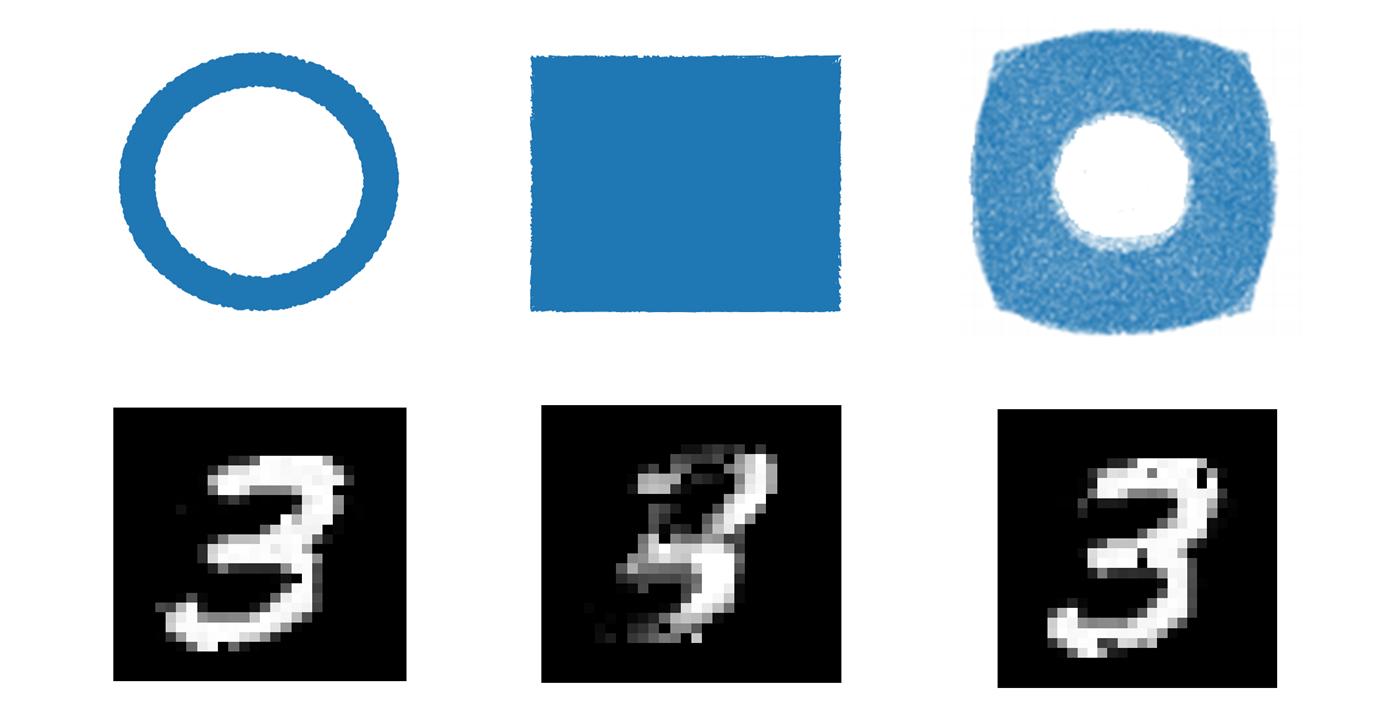}
\centering
\caption{Qualitative results of \baby in 2D setting. The first and second columns are input distributions, and the third column is the barycenter learned by \baby. }
\label{Fig2}
\end{figure}

Figure \ref{Fig2} shows the qualitative performance of our method on 2D examples. Each example is represented in a row. The first and second columns are input distributions, and the third column is the approximated barycenter by using our method. In the first example, we set the variational distribution $\nu'$ to be a Gaussian mixture with 30 components. In the second example, we set $\nu'$ to be a Gaussian mixture with 20 components. We can find that \baby learns the barycenter qualitatively well. For comparison with CRWB and SCWB, see \citep[Figure 1]{li2020} and \citep[Figure 4]{fan2020}.
\subsection{Evaluation on Multivariate Gaussians}
Though \cite{korotin2022} constructs a dataset for quantitative evaluation of barycenter methods, this dataset may be only suitable for methods using generative models in high-dimensional settings (e.g., $\ge 16$). Therefore, for quantitative evaluation we compute the Wasserstein barycenters of $N$ multivariate Gaussians $\mathcal{N}(m_n,\Sigma_n)$ in dimension $D$ with mean $m_n$ and non-diagonal covariance matrix $\Sigma_n$. When the input distributions are multivariate Gaussians, the Wasserstein barycenter is always a Gaussian $\mathcal{N}(m_*,\Sigma_*)$, where $m_*=\sum\nolimits_{n = 1}^N{\alpha_n m_n}$ and $\Sigma_*=\sum\nolimits_{n = 1}^N{\alpha_n (\Sigma_*^{\frac{1}{2}}\Sigma_n\Sigma_*^{\frac{1}{2}})^{\frac{1}{2}}}$ computed through an efficient fixed-point algorithm \citep{alvarez2016}.

To evaluate the performance of these methods, we use the \textbf{B}ures-\textbf{W}asserstein \textbf{U}nexplained \textbf{V}ariance \textbf{P}ercentage (BW$_2^2$-UVP) to measure whether the approximation of the Wasserstein barycenter, denoted by $\nu$ with mean $m_{\nu}$ and covariance $\Sigma_{{\nu}}$, approaches the true one $\tilde {\nu}$ with mean $m_{\tilde \nu}$ and covariance $\Sigma_{{\tilde \nu}}$ \citep{korotin2021}:
\begin{equation*}
    \rm {BW_2^2-UVP}(\nu,\tilde{\nu}) \buildrel \Delta \over=100\frac{\rm {BW^2_2}(\nu,\tilde{\nu})}{\frac{1}{2}\rm {Var}(\tilde{\nu})}\%,
\end{equation*}
where $\rm {BW^2_2} (\nu,{\tilde{\nu}})$ equals
\begin{equation*}
\label{EqCFS}
\frac{1}{2}{ \| {m _\nu} - {m_{\tilde{\nu}}}\| ^2 + \frac{1}{2}{\rm{Tr}}\left({\Sigma _\nu} + {\Sigma_{\tilde{\nu}}} - 2{(\Sigma_{\nu}^{{1 \mathord{\left/{\vphantom {2 1}} \right.
 \kern-\nulldelimiterspace} 2}}{\Sigma _{\tilde{\nu}}}\Sigma _{\nu}^{{1 \mathord{\left/{\vphantom {1 2}} \right.\kern-\nulldelimiterspace} 2}})^{{1 \mathord{\left/{\vphantom {1 2}} \right.\kern-\nulldelimiterspace} 2}}}\right)}.
\end{equation*}
Here $\rm{Tr}(\cdot)$ denotes the trace of a matrix. When $\rm {BW_2^2-UVP} \approx 0 \%$, $\nu$ is a good approximation of ${\tilde{\nu}}$. In the case of $\rm {BW_2^2-UVP} \geq 100 \%$, the approximation $\nu$ is undesirable.

\begin{figure}[hb]
\includegraphics[width=0.45\textwidth,height=0.5\textwidth]{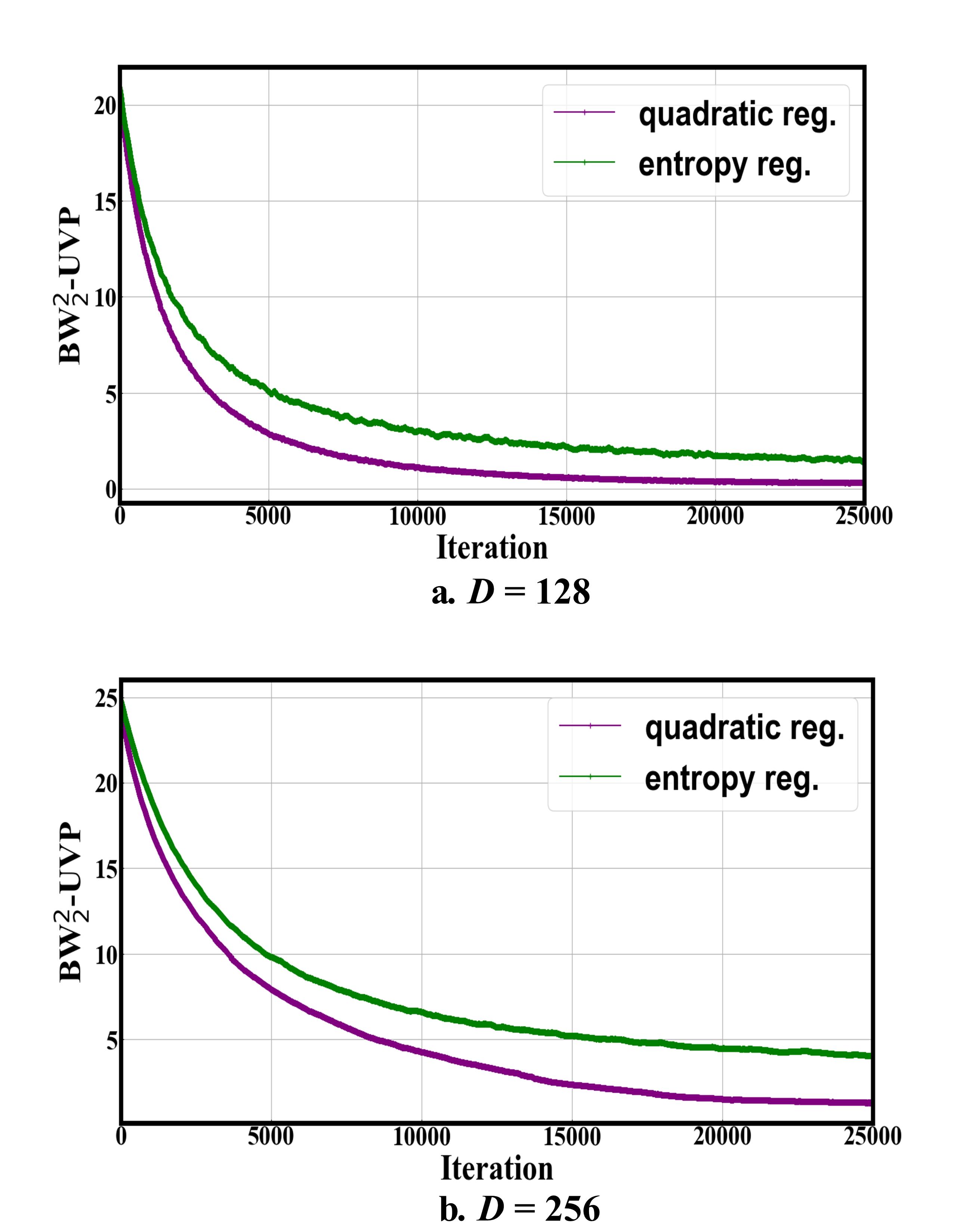}
\centering
\caption{Convergence curves of \baby with quadratic and entropy regularization at $D$=128 and 256. }
\label{Fig3}
\end{figure}

The results of 3 randomly generated multivariate Gaussians in varying dimensions $D$ are shown in Table 2. We can find that the estimation errors of \baby are lower than those of other methods in all cases. These results imply that \baby can make good approximations of continuous Wasserstein barycenters, in particular for high-dimensional cases. To further demonstrate the effectiveness of $c$-cyclical monotonicity regularization, we give the result of the ablative version of \baby, namely VWB, which computes the barycenters based on the original regularized objective (\ref{rwb}) without this regularization. We see that \baby is significantly better than VWB. This result directly indicates the positive impact of $c$-cyclical monotonicity. On the other hand, VWB outperforms CRWB, which shows the effectiveness of the variational distribution. Because the main difference between VWB and CRWB is that VWB introduces a variational distribution to represent the barycenter instead of a fixed prior in CRWB.
\begin{table*}
\renewcommand\arraystretch{1.4}
\label{Table2}
\begin{centering}
\begin{tabular}{c|c|c|c|c|c|c|c|c}
\Xhline{1.2pt}
Method & $ D=2$ &4 &8 &16 &32 &64 &128 &256 \\
\hline
\baby  \textbf{(Ours)} &\textbf{0.02}&\textbf{0.06}& \textbf{0.11}&\textbf{0.13}&\textbf{0.17}& \textbf{ 0.20 } & \textbf{0.40}& \textbf{1.10}\\
VWB & 0.04 &0.11 &0.21 &0.24 &0.30 & 0.77 & 5.01 & 14.65\\
CRWB \citep{li2020} &0.12 &0.62&0.84&9.58&25.81& 26.31&51.75 & \textgreater{100}\\
CWB \citep{korotin2021} &0.06 &0.10 &0.17 &0.20 &0.21 &0.25 &1.92 &4.81\\
SCWB \citep{fan2020} & 0.03& 0.12& 0.16& 0.21& 0.40& 0.67& 1.42 &3.41\\
\Xhline{1.2pt}
\end{tabular}
\caption{Numerical results of the $\rm{BW}^2_2-\rm{UVP}$ error for estimating barycenters of Gaussians of varying dimensions. Smaller is better. The second row lists the results of the ablative version of \baby without the $c$-cyclical monotonicity regularization (CMR).}
\end{centering}
\end{table*}
\paragraph{Convergence.}
Figure \ref{Fig3} shows the convergence curves of \baby with quadratic and entropy regularization at $D$=128 and 256 within 25000 iterations. Though the optimization over $\nu'$ is not convex, we can see that \baby can converge to a local optimum. Besides, the empirical results show that quadratic regularization is more stable than entropy regularization, since there is no exponential term causing overflow.
\subsection{Evaluation on Real Data}
We now evaluate \baby in real-world applications. We apply the Wasserstein barycenter to aggregate posterior distributions of subsets in a large data set to approximate the true posterior of the full data. Following \citep{li2020}, we use Poisson regression for the task of predicting the hourly number of bike rentals using features such as the day of the week and weather conditions \footnote{http://archive.ics.uci.edu/ml/datasets/Bike+Sharing+Dataset}. We consider the posterior distribution on the 8-dimensional regression coefficients for the Poisson model. We firstly randomly split the full data into 5 subsets with equal-size and get $10^5$ samples from each subset posterior using the Stan library \citep{carpenter2017}. Then we compute the Wasserstein barycenter of these subset posterior distributions by all methods. In our method, we set $\nu'$ to be a Gaussian. To evaluate the performance of these methods, we compute the BW$_2^2$-UVP between the approximated barycenter and the full data posterior.

The results are shown in Table 3. All methods approach the true Wasserstein barycenter well since the values of $\rm {BW}_2^2-\rm{UVP} \textless 1\%$. The performance of \baby is better than CRWB, CWB and SCWB. This further indicates that \baby is an effective alternative to compute continuous Wasserstein barycenters.

\begin{table}[t]
\renewcommand\arraystretch{1.4}
\centering
\begin{tabular}{c|c}
\Xhline{1.2pt}
Method &  $\rm {BW}_2^2-\rm{UVP}$\\
\hline
 CRWB \citep{li2020} &0.96\\
 CWB \citep{korotin2021}& 0.13\\
 SCWB \citep{fan2020}& 0.17\\
 \baby \textbf{(Ours)}&\textbf{0.10} \\
\Xhline{1.2pt}
\end{tabular}
\caption{Empirical results on subset posterior aggregation. The $\rm {BW}_2^2-\rm{UVP}$ is computed for comparison by CRWB, CWB, SCWB and \baby.}
\label{Table3}
\end{table}

\section{Conclusion}
We develop a novel \baby method to compute the Wasserstein barycenters of continuous distributions. The main idea is to explore $c$-cyclical monotone condition to construct a new regularized objective and introduce a variational distribution to represent the barycenter. The objective can be solved by stochastic optimization only requiring samples access to the input distributions, where variational parameters are optimized to find the closest distribution as the approximation of the barycenter. The limitation of our method is that we require to adjust the regularization parameter in different applications. One future direction is to find a way to solve this issue dynamically.

\section{Acknowledgments}
We would like to acknowledge support for this project from the National Natural Science Foundation of China (NSFC) (No. 62006094, No.62276113) and National Key R\&D Program of China (No.2021ZD0112501, No.2021ZD0112502). Besides, we thank the inspiring discussions with Lingxiao Li from MIT University and the insightful lectures about the Optimal Transport theory from Xiangfen David Gu from Stony Brook University.

\bibliography{aaai23}

\end{document}